\documentclass{article}

\PassOptionsToPackage{numbers, compress}{natbib}
\usepackage[preprint]{neurips_2021}

\usepackage[utf8]{inputenc} 
\usepackage[T1]{fontenc}    
\usepackage{hyperref}       
\usepackage{url}            
\usepackage{booktabs}       
\usepackage{amsfonts}       
\usepackage{nicefrac}       
\usepackage{microtype}      
\usepackage{xcolor}         
\usepackage[pdftex]{graphicx}

\title{Distilling Self-Knowledge From Contrastive Links to Classify Graph Nodes Without Passing Messages}

\author{%
  Yi Luo\\
  School of Computer Science and Engineering\\
  University of Electronic Science and Technology of China\\
  Chengdu 611731, China\\
  \texttt{cf020031308@163.com}\\
  \And
  Aiguo Chen\thanks{Corresponding Author. E-mail: agchen@uestc.edu.cn}\\
  School of Computer Science and Engineering\\
  University of Electronic Science and Technology of China\\
  Chengdu 611731, China\\
  \texttt{agchen@uestc.edu.cn}
  \And
  Ke Yan\\
  School of Computer Science and Engineering\\
  University of Electronic Science and Technology of China\\
  Chengdu 611731, China\\
  \texttt{kyan@uestc.edu.cn}
  \And
  Ling Tian\\
  School of Computer Science and Engineering\\
  University of Electronic Science and Technology of China\\
  Chengdu 611731, China\\
  \texttt{lingtian@uestc.edu.cn}\\
}

\begin{document}

\maketitle

\begin{abstract}
    Nowadays, Graph Neural Networks (GNNs) following the Message Passing paradigm become the dominant way to learn on graphic data.
    Models in this paradigm have to spend extra space to look up adjacent nodes with adjacency matrices and extra time to aggregate multiple messages from adjacent nodes.
    To address this issue, we develop a method called LinkDist that distils self-knowledge from connected node pairs into a Multi-Layer Perceptron (MLP) without the need to aggregate messages.
    Experiment with 8 real-world datasets shows the MLP derived from LinkDist can predict the label of a node without knowing its adjacencies but achieve comparable accuracy against GNNs in the contexts of semi- and full-supervised node classification.
    Moreover, LinkDist benefits from its Non-Message Passing paradigm that we can also distil self-knowledge from arbitrarily sampled node pairs in a contrastive way to further boost the performance of LinkDist. 
\end{abstract}

\section{Introduction}
\label{sec:introduction}

Graphs are universal representations for many real-world data containing pairwise relationship.
For example, academic papers and their citation relationship can construct a graph, where the papers are represented as nodes with their encoded abstracts as node features.
If one paper cites another, their nodes are connected in that graph.
Graphs like this are usually called citation networks~\citep{DBLP:journals/aim/SenNBGGE08}.
Likewise, we have co-purchase networks~\citep{DBLP:conf/sigir/McAuleyTSH15}, co-authorship networks~\citep{DBLP:journals/qss/WangSHWDK20}, protein-protein interaction networks (PPI), post-post co-comment networks~\citep{DBLP:conf/nips/HamiltonYL17}, and so on~\citep{wang2019dgl}.

Due to the wide application of graphs, researchers propose more and more methods to mine information from graphs.
A well-investigated family of methods is Graph Neural Networks (GNNs)~\citep{DBLP:journals/tnn/WuPCLZY21}.
They combine node features and the structural information of graphs to study the representations of nodes for downstream tasks, such as Node Classification where the goal is to classify nodes by assigning proper labels to them.

Many Graph Neural Networks follow the Message Passing paradigm~\citep{DBLP:journals/corr/GilmerSRVD17} which contains three steps:

\begin{enumerate}
    \item Each node passes its representation to its adjacent nodes
    \item Each node aggregates the representations it receives
    \item Each node transforms the aggregated information to get its newest representation
\end{enumerate}

For example, a layer in Graph Convolutional Network (GCN)~\citep{DBLP:journals/corr/KipfW16} aggregates each node's received messages by computing their average and transforms the averaged messages with a linear transformation.

This paradigm applies Neural Networks to graphs and makes learning on graphs flexible and powerful.
However, it still has some drawbacks.

\begin{itemize}
    \item Predicting a single node's label has to aggregate information from multiple adjacent nodes, making the computation heavy and costing extra space to look up adjacent nodes (usually with the adjacency matrix).
    \item Passing messages is not easy in practice.
        For batching purposes, we have to group nodes by their degrees (as the number of their adjacent nodes) and invokes message passing group by group if the graph is too large to manipulate its adjacency matrix.
    \item Relying on information from adjacent nodes takes the graph time and effort to propagate messages onto all nodes when new nodes or new edges are inserted.
\end{itemize}

One way to rectify some of the aforementioned issues is Knowledge Distillation~\citep{DBLP:journals/corr/abs-2006-05525}.
Knowledge Distillation is to transfer the knowledge from a well-trained teacher network to a smaller student network for deployment purposes.
For example, when the spaces and computing resources on devices are too limited to store adjacency matrices or aggregate multiple messages timely, we can distil the knowledge from a trained GCN into a Multi-Layer Perceptron (MLP)~\citep{DBLP:books/lib/Bishop07} and deploy the MLP alone onto those devices.
The deployed MLP can predict without the need to aggregate information from adjacent nodes, thus it can be light-weighted and fast. 
However, when graphs change, the MLP still has to study the updated knowledge about every single node from the GCN again.
This makes Knowledge Distillation hard to apply to dynamic graphs.

However, we observe that the MLP distilled from a trained GCN can sometimes outperform the original GCN, as our experiment in Section~\ref{sec:experiment} shows.
It suggests that Message Passing may not be essential if we can find a way to distil the structural knowledge into MLPs.
Inspired by this, we propose to distil knowledge from node pairs.
Specifically, we predict the label of a node by its features with an MLP and predict the label of its adjacent node with a forked layer.
In the meantime, we maximize the agreement of these two predictions by distilling their knowledge into each other.
After that, we omit the forked layer and get an MLP that contains distilled structural knowledge.
This method, named \emph{LinkDist}, is light-weighted, fast, and easy to implement.
When graphs change, LinkDist adjusts its parameters instead of propagating messages to all nodes.
While addressing all the three challenges in Message Passing, the Node Classification experiment in Section~\ref{sec:experiment} shows that the MLPs derived from LinkDist can predict with comparable accuracy against GNNs.

\section{Related Works}

\paragraph{Self-Knowledge Distillation}
Besides the Teacher-Student scheme of Knowledge Distillation, a similar technique known as Self-Knowledge Distillation~\citep{1905.08094v1} regularizes a model by distilling its own knowledge without teacher models.
An effective way to achieve this is to induce predictions that are consistent with relevant data such as the universal label distribution of same-labelled nodes~\citep{DBLP:conf/cvpr/YunPLS20}.
Our method LinkDist works similarly.
It distils its own knowledge by predicting the label of a node consistently with both the node itself and its adjacent nodes.

\paragraph{Contrastive Learning}
Contrastive Learning~\citep{DBLP:journals/corr/abs-2006-08218} is a technique that learns to distinguish data points without labels by pulling similar points together and pushing dissimilar points away.
Likewise, we propose a contrastive way to train LinkDist.
While we distil self-knowledge by minimizing the difference of predictions made with two connected nodes, we also maximize the difference of predictions made with two arbitrarily sampled nodes.
Besides, different from Contrastive Learning that learns without labels, we pull predictions made with connected nodes to the ground-truth labels and push predictions made with arbitrarily sampled node pairs away from them.

\section{Our Method: LinkDist}

In this section, we describe LinkDist that leverages a forked Multi-Layer Perceptron to produce predictions on nodes both from their features and from their locality.
It is trained by iterating on the edge set and distilling self-knowledge from links to induce consistent predictions.
We then propose a method of contrastive training to distil knowledge from links that do not exist.
After training, we evaluate LinkDist in two modes.
One is faster, and one is more accurate.

\subsection{Forked Multi-Layer Perceptron}

Our objective is to learn a mapping $M: x \rightarrow (z, s)$ that receives the features $x$ of a node $v$ and predicts both the node's label $y$ and the label of its adjacent node.
In graphic data, aggregating messages from adjacent nodes can help a node to determine its label.
Vice versa, it is reasonable to partly recover the label of adjacent node with features of a node.

The mapping $M: x \rightarrow (z, s)$ can be implemented as a forked Multi-Layer Perceptron with hidden layers, an output layer, and an inference layer.
We feed a node $v$'s features $x$ into the hidden layers, encoding them into a hidden representation $h$.
Then, the output layer and the inference layer use the hidden representation $h$ to produce the logits $z, s$ of label distributions respectively, where $z$ is the approximation of $v$'s label $y$ and $s$ is the possible label distribution of $v$'s adjacent node.

%

\subsection{The Training of LinkDist}
\label{ssec:train}

\begin{figure}
  \includegraphics[width=\linewidth]{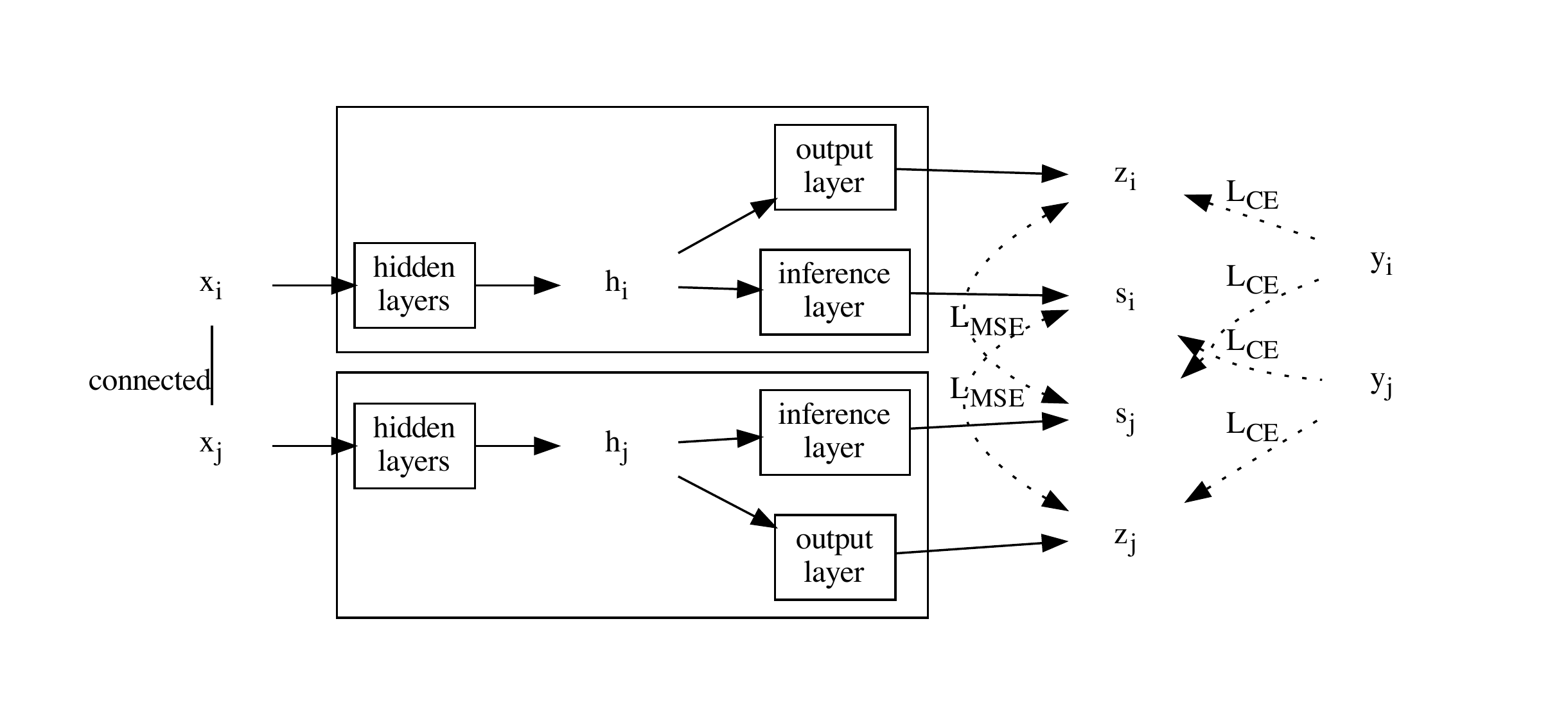}
  \centering
  \caption{
      The training of LinkDist.
      The features $x_i, x_j$ of two connected nodes $(v_i, v_j)$ are fed into the forked MLP with hidden layers, an output layer, and an inference layer.
      The forked MLP produces $z_i, s_i, z_j, s_j$ as approximations to the label of $v_i$, the label of $v_i$'s adjacent node, the label of $v_j$ and the label of $v_j$'s adjacent node.
      They are supervised by the ground-truth labels $y_i, y_j$ of $v_i, v_j$ if are available.
      $s_i$ and $s_j$ contain dark knowledge of labelled nodes and additional knowledge of unlabelled nodes.
      So we distil knowledge from $s_i$ into $z_j$ and from $s_j$ into $z_i$.
  }
  \label{fig:arch}
\end{figure}

We illustrate the training of LinkDist in Figure~\ref{fig:arch}.
In the training phase, we iterate on the edge set $E = \{ (v_i, v_j), \cdots \}$ and feed the features $x_i, x_j$ of each node pair $(v_i, v_j)$ into the forked MLP, getting the approximations $z_i, z_j$ of their labels $y_i, y_j$ and the approximations $s_i, s_j$ of their adjacent nodes' labels.

\paragraph{Ground-Truth Label Supervision}
If the node $v_i$ is assigned with a ground-truth label $y_i$, we supervise $z_i$ by minimizing its Cross-Entropy loss with $y_i$.
Since $v_i$ and $v_j$ are each other's adjancet nodes, the label $y_i$ is also the possible distribution of $s_j$, so we also supervise $s_j$ with $y_i$.
Likewise, if the node $v_j$ is assigned with a ground-truth label $y_j$, we supervise both $z_j$ and $s_i$ with $y_j$.

In addition, the label distribution of the edge set that we iterate on is different from that of the node set.
To predict labels of nodes in the node set, we assign weights $y_n / y_e$ to labels when computing the Cross-Entropy loss $L_{CE}$, where $y_n$ is the label distribution in the training set and $y_e$ is the label distribution in the edge set with unlabelled nodes masked.

\paragraph{Link Distillation}
For each node, we have two predictions of its label from different sources.
One is from its features.
Another one is from its adjacent node.
By minimizing their difference $L_{MSE}$ measured with Mean Square Error, we can distil knowledge from both of them into each other, maximizing the agreement of predicting the label of a node by its feature information and by the structural knowledge from its adjacent node. 

In total, we optimize $L_{CE} + \alpha \cdot L_{MSE}$ to cast both ground-truth label supervision and link distillation.
By default, we set the hyperparameter $\alpha = 1 - n_e / e$ where $n_e$ is the times that labelled nodes appear in the edge set.
In other words, the more labelled nodes containing supervising information we have, the less important the knowledge inferred from adjacent nodes is.

\subsection{Contrastive Training With Negative Links}

\begin{figure}
  \includegraphics[width=\linewidth]{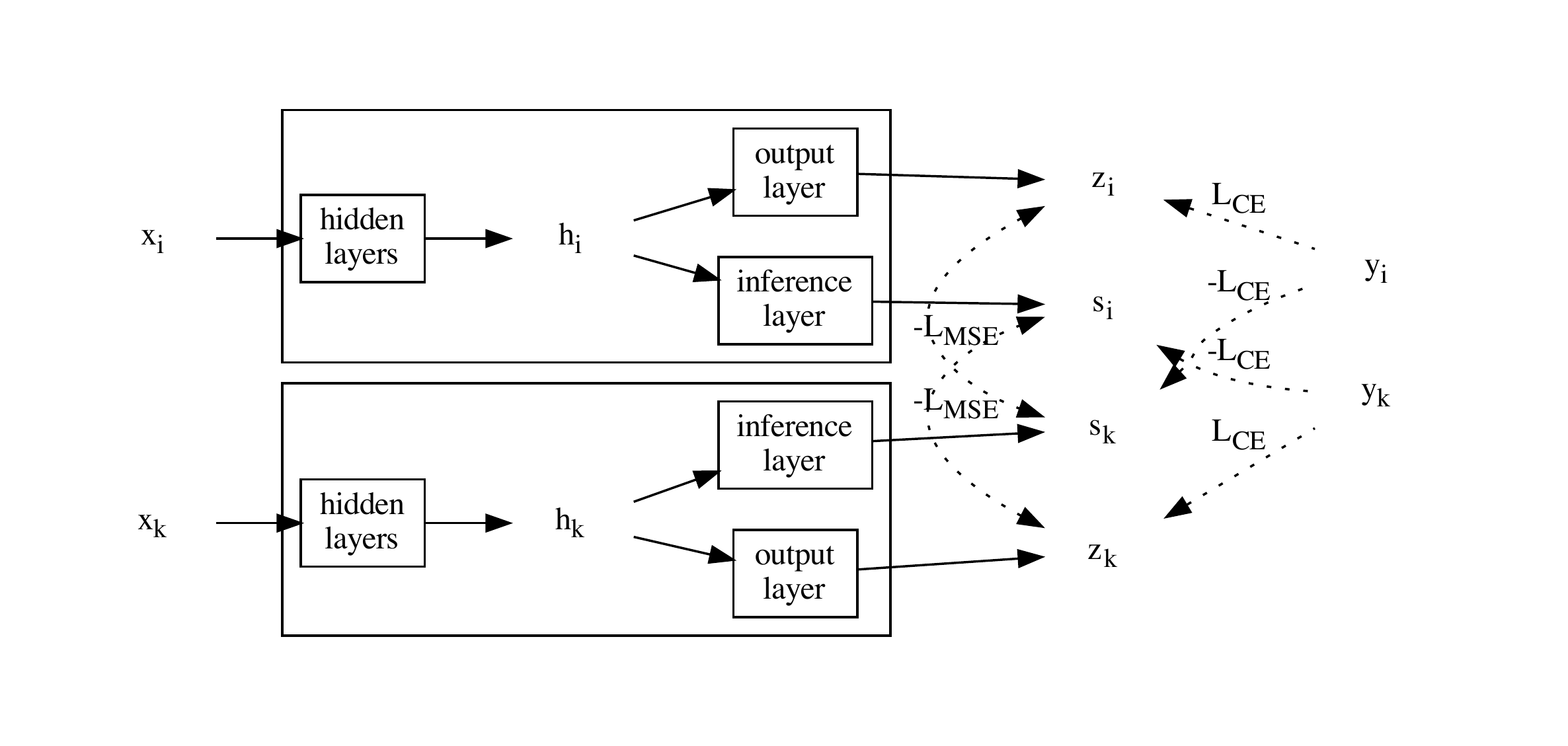}
  \centering
  \caption{
      Contrastive training with negative links.
      The features $x_i, x_k$ of two arbitrarily sampled nodes $(v_i, v_k)$ are fed into the forked MLP\@.
      The forked MLP produces $z_i, s_i, z_k, s_k$ as approximations to the label of $v_i$, the label of $v_i$'s adjacent node, the label of $v_k$ and the label of $v_k$'s adjacent node.
      $z_i, z_k$ are supervised by the ground-truth labels $y_i, y_k$ of $v_i, v_j$ if are available.
      Since the node $v_k$ is not likely to connect $v_i$, we push $s_i$ away from $z_k$ and $y_k$ by maximizing their differences measured by Mean Square Error and Cross-Entropy respectively.
      Likewise, we push $s_k$ away from $z_i$ and $y_i$.
  }
  \label{fig:neg}
\end{figure}

To leverage node pairs that are not connected, we propose a method of contrastive training as illustrated in Figure~\ref{fig:neg}.

We sample arbitrary pairs of nodes to construct negative links~\citep{DBLP:conf/kdd/GroverL16} and feed the features $x_i, x_k$ of each node pair $(v_i, v_k)$ into the forked MLP, getting the approximations $z_i, z_k$ of their labels $y_i, y_k$ and the approximations $s_i, s_k$ of their adjacent nodes' labels.

While we supervise $z_i$ with $y_i$ and supervise $z_k$ with $y_k$ as we do in the previous section~\ref{ssec:train}, we maximize the Cross-Entropy of $s_i, z_k$ and the Cross-Entropy of $s_k, z_i$.
This is because node $v_k$ is arbitrarily sampled that it is not likely to connect $v_i$.
Therefore, the approximation $z_i$ of node $v_i$'s label is probably different from the approximation $s_k$ of labels assigned to node $v_k$'s adjacent nodes and $z_k$ is probably different from $s_i$.
Likewise, we maximize the MSE of $z_i, s_k$ and the MSE of $z_k, s_i$ to distil self-knowledge from negative links.

This contrastive training method has to be adopted with the training process described in the previous section~\ref{ssec:train}.
In this work, we name the contrastively trained model \emph{CoLinkDist}.

\subsection{Two Evaluating Modes of LinkDist}

The forked MLP in LinkDist outputs the label distribution $z$ of the node with its output layer and the label distribution $s$ of the node's adjacent node with its inference layer.
When omitting the inference layer and its output $s$, the forked MLP becomes an MLP, but with structural knowledge distilled from links.
The MLPs derived from LinkDist and CoLinkDist, named \emph{LinkDistMLP} and \emph{CoLinkDistMLP}, can classify graph nodes fast without aggregating messages from their adjacent nodes, thus they can be light-weighted to deploy without adjacency matrices or other nodes' features.
Moreover, the experiment in Section~\ref{sec:experiment} shows the accuracy of LinkDistMLP outperforms that of vanilla MLPs by large margins, approaching the accuracy of GCNs.

When adjacency matrices and features of adjacent nodes are available, LinkDist can also utilize Message Passing to boost its accuracy.
It combines the prediction from the node $v_i$ itself and the predictions from its adjacent nodes to estimate its possible labe $\hat y_i$ with:
\[\hat y_i = z_i + \alpha \cdot \sum\limits_{v_j \in N(v_i)} s_j\]
where $N(v_i)$ is the set of nodes that are adjacent to node $v_i$.

\section{Experiment: Node Classification}
\label{sec:experiment}

In this section, we conduct the Node Classification experiment on 8 real-world datasets.
The code to reproduce this work is published at \url{https://github.com/cf020031308/LinkDist}.

\subsection{Datasets}

\begin{table}
  \caption{Summary of the 8 real-world datasets}
  \label{tbl:datasets}
  \centering
  \begin{tabular}{l|rrrrr}
    \toprule
      Dataset & \#Nodes & \#Features & \#Classes & \#Edges & Average Degree \\
    \midrule
      Cora             & 2708  & 1433 & 7  & 5278   & 3.90  \\
      Citesee          & 3327  & 3703 & 6  & 4552   & 2.77  \\
      Pubmed           & 19717 & 500  & 3  & 44324  & 4.50  \\
      Cora Full        & 19793 & 8710 & 70 & 63421  & 6.41  \\
      Amazon Photo     & 7650  & 745  & 8  & 119081 & 31.13 \\
      Amazon Computer  & 13752 & 767  & 10 & 245861 & 35.76 \\
      Coauthor CS      & 18333 & 6805 & 15 & 81894  & 8.93  \\
      Coauthor Physics & 34493 & 8415 & 5  & 247962 & 14.38 \\
    \bottomrule
  \end{tabular}
\end{table}

To fully evaluate our method, we experiment with 8 real-world datasets including 4 citation networks, 2 co-purchase networks and 2 co-authorship networks.
They are Cora, Citeseer, Pubmed~\citep{DBLP:journals/aim/SenNBGGE08}, Cora Full~\citep{DBLP:conf/iclr/BojchevskiG18}, Amazon Photo, Amazon Computer~\citep{DBLP:conf/sigir/McAuleyTSH15}, Coauthor CS and Coauthor Physics~\citep{DBLP:journals/qss/WangSHWDK20}.
These datasets are very different in many respects.
We summarize them in Table~\ref{tbl:datasets}.

\paragraph{Semi- and Full-Supervised}

Among the 8 datasets, Cora, Citeseer and Pubmed are distributed with their predefined dataset splits.
In detail, in each of these 3 datasets, the training set contains 20 nodes from each class, the validation set contains 500 nodes and the testing set contains 1000 nodes.
On Cora, Citeseer and Pubmed, we accordingly use these splits to do the semi-supervised Node Classification experiment.
On other datasets, we split nodes into training sets, validation sets and testing sets exactly like on Cora, Citeseer and Pubmed.
In particular, we randomly select at most 20 nodes from each class to construct the training set, 500 nodes to construct the validation set, and 1000 nodes to construct the testing set.

In addition to this semi-supervised setting, we further split each dataset with the full-supervised setting.
Specifically, we randomly select 60\% of the nodes in a dataset to construct its training set.
Then, we equally divide the rest nodes into a validation set and a testing set.

\paragraph{Transductive and Inductive}

In the semi-supervised Node Classification, we experiment with both the transductive setting and the inductive setting.
With the transductive setting, methods use all nodes features and connections in the training phase.
With the inductive setting, nodes in the two evaluation sets (the validation set and the testing set) are masked in the training phase.
So are connections that have endpoints in the evaluation sets.
This setting meets practical scenarios where graphs are consistently expanding with new nodes~\citep{DBLP:conf/iclr/ChenMX18}.

In the full-supervised Node Classification, the majority of nodes are either in the evaluation sets or connected with nodes from the evaluation sets.
So we experiment with only the transductive setting.

\subsection{Baselines}

We implement 3 methods as baselines.

\paragraph{Graph Convolutional Network (GCN)}
LinkDist working in Message Passing paradigm can utilize all the information that most GNNs use.
So we compare LinkDist against GCN to examine its applicability in common situations.
Due to the inductive setting, we implement GCN with its normalized adjacency matrix as $(D + I)^{-1}(A + I)$ instead of $(D + I)^{-\frac12}(A + I)(D + I)^{-\frac12}$.
$I$ is the identity matrix, $A$ is the adjacency matrix, and $D$ is a diagonal matrix where each diagonal entry $d_i$ is the degree of node $v_i$.

\paragraph{Multi-Layer Perceptron (MLP)}
In Non-Message Passing mode, LinkDistMLP works just like an MLP\@.
However, with knowledge distilled from links, LinkDistMLP should go beyond it.
So we place an MLP here to act as the lower limit of the LinkDistMLP.

\paragraph{MLP distilled from GCN (GCN2MLP)}
We teach an MLP the soft label logits of all nodes outside the evaluation sets predicted by a trained GCN\@.
The distilled MLP has an identical structure to both the MLP and the LinkDistMLP but inherits the structural knowledge from GCNs.
We think it is a proper benchmark for LinkDistMLP to measure how far it is from `the perfection'.

All implementations of MLP and GCN have 3 full-connected layers.
The LinkDist has 2 hidden layers, 1 output layer and 1 inference layer, all implemented as full-connected layers.
Hidden representations within all layers have the identical size 256.
There are 1 Batch Normalization~\citep{DBLP:conf/icml/IoffeS15}, 1 Layer Normalization~\citep{DBLP:journals/corr/BaKH16}, 1 Dropout~\citep{DBLP:journals/corr/abs-1207-0580} with 0.5 possibilities to erase a position and 1 LeakyReLU~\citep{DBLP:journals/corr/XuWCL15} activator between every two layers.

All the methods can be divided into two groups by whether a method leverages information from adjacent nodes in the evaluation phase.
MLP, GCN2MLP, LinkDistMLP, and CoLinkDistMLP involve no Message Passing, whereas GCN, LinkDist, and CoLinkDist do.

\subsection{Results}
\label{ssec:results}

We use Adam~\citep{DBLP:journals/corr/KingmaB14} with a learning rate of 0.01 to optimize parameters.
The GCN is trained in full batches.
The MLP and the LinkDist are trained in mini-batches with the batch size set to 1024.
Since the MLP and the GCN iterates on the node set while the LinkDist iterate on the edge set, we train the former two models for 200 epochs but train the LinkDist for about $200 / \tilde d$ epochs where $\tilde d$ is the average degree of the graph.

\begin{table}
  \caption{
      Accuracy Scores (\%) in the \textbf{semi-supervised} and \textbf{inductive} Node Classification.
      Methods are divided into two groups by whether the method aggregates messages from adjacent nodes in the evaluation phase.
      We bold the highest accuracy for every dataset in each group.}
  \label{tbl:inductive}
  \centering
  \setlength{\tabcolsep}{1.4mm}{
  \begin{tabular}{lrrrrrrrr}
    \toprule
      Method & Cora & Citesee & Pubmed & Cora & Amazon & Amazon & Coauthor & Coauthor \\
      & & & & Full & Photo & Computer & CS & Physics \\
    \midrule
MLP & 56.28          & 54.74          & 70.33          & 41.91          & 75.89          & 66.14          & 90.11          & 89.53 \\
GCN2MLP & 67.28          & 64.19          & \textbf{76.72} & \textbf{53.59} & \textbf{87.83} & \textbf{79.36} & \textbf{93.13} & \textbf{93.20} \\
LinkDistMLP & 66.38          & 67.85          & 72.78          & 47.15          & 86.16          & 76.44          & 89.20          & 90.68 \\
CoLinkDistMLP & \textbf{68.93} & \textbf{69.85} & 74.32          & 49.88          & 86.34          & 76.48          & 90.16          & 91.38 \\
    \midrule
GCN & 77.12          & 62.98          & 73.00          & 56.88          & 87.63          & 79.61          & 88.21          & 92.67 \\
LinkDist & 75.72          & 71.19          & \textbf{74.84} & 56.09          & 89.67          & \textbf{82.36} & 91.64          & 92.84 \\
CoLinkDist & \textbf{77.14} & \textbf{71.74} & 74.77          & \textbf{57.33} & \textbf{90.56} & 81.93          & \textbf{92.20} & \textbf{93.04} \\
    \bottomrule
  \end{tabular}}
\end{table}

\begin{table}
  \caption{
      Accuracy Scores (\%) in the \textbf{semi-supervised} and \textbf{transductive} Node Classification.
      Methods are divided into two groups by whether the method aggregates messages from adjacent nodes in the evaluation phase.
      We bold the highest accuracy for every dataset in each group.}
  \label{tbl:transductive}
  \centering
  \setlength{\tabcolsep}{1.4mm}{
  \begin{tabular}{lrrrrrrrr}
    \toprule
      Method & Cora & Citesee & Pubmed & Cora & Amazon & Amazon & Coauthor & Coauthor \\
      & & & & Full & Photo & Computer & CS & Physics \\
    \midrule
MLP           & 56.28          & 54.74          & 70.33          & 41.91          & 75.89          & 66.14          & 90.11          & 89.53 \\
GCN2MLP       & 67.61          & 63.29          & \textbf{77.87} & \textbf{54.15} & 87.11          & 78.21          & \textbf{92.49} & \textbf{93.42} \\
LinkDistMLP   & 80.79          & 70.26          & 72.41          & 51.78          & 88.19          & 78.23          & 89.83          & 90.72 \\
CoLinkDistMLP & \textbf{81.19} & \textbf{70.96} & 75.41          & 53.43          & \textbf{88.90} & \textbf{78.77} & 90.66          & 91.63 \\
    \midrule
GCN        & 76.47          & 63.11          & 73.99          & \textbf{57.58} & 87.00          & 77.95          & 87.93          & \textbf{92.79} \\
LinkDist   & 81.05          & 70.27          & 74.06          & 55.87          & 90.14          & \textbf{82.65} & 90.94          & 92.06 \\
CoLinkDist & \textbf{81.39} & \textbf{70.79} & \textbf{75.64} & 57.05          & \textbf{90.54} & 82.53          & \textbf{91.88} & 92.74 \\
    \bottomrule
  \end{tabular}}
\end{table}

\begin{table}
  \caption{
      Accuracy Scores (\%) in the \textbf{full-supervised} and \textbf{transductive} Node Classification.
      Methods are divided into two groups by whether the method aggregates messages from adjacent nodes in the evaluation phase.
      We bold the highest accuracy for every dataset in each group.}
  \label{tbl:full-supervised}
  \centering
  \setlength{\tabcolsep}{1.4mm}{
  \begin{tabular}{lrrrrrrrr}
    \toprule
      Method & Cora & Citesee & Pubmed & Cora & Amazon & Amazon & Coauthor & Coauthor \\
      & & & & Full & Photo & Computer & CS & Physics \\
    \midrule
      MLP           & 74.71          & 71.95          & 87.92          & 62.03          & 91.75          & 85.08          & 95.35          & 96.57 \\
      GCN2MLP       & 75.95          & 72.93          & 86.25          & 62.39          & 91.10          & 85.41          & 95.29          & 96.46 \\
      LinkDistMLP   & \textbf{87.58} & 75.25          & 88.79          & 69.53          & 93.83          & \textbf{89.44} & 95.68          & \textbf{96.91} \\
      CoLinkDistMLP & 87.54          & \textbf{75.77} & \textbf{89.53} & \textbf{69.83} & \textbf{94.12} & 88.85          & \textbf{95.74} & 96.87 \\
    \midrule
      GCN           & 86.03          & 73.37          & 84.79          & 68.97          & 93.09          & \textbf{90.08} & 92.27          & 95.96 \\
      LinkDist      & \textbf{88.24} & 74.72          & 88.86          & 69.87          & 93.75          & 89.49          & 95.66          & 96.87 \\
      CoLinkDist    & 87.89          & \textbf{75.79} & \textbf{89.58} & \textbf{70.32} & \textbf{94.36} & 89.42          & \textbf{95.80} & \textbf{97.05} \\
    \bottomrule
  \end{tabular}}
\end{table}

After every epoch of training, we evaluate the method on the two evaluation sets, producing a pair of accuracy scores.
After running, we get the testing set accuracy score paired with the highest validation set accuracy score as the score of the evaluated method.
We run every method on every dataset 10 times and record the average score in Table~\ref{tbl:inductive}, Table~\ref{tbl:transductive}, and Table~\ref{tbl:full-supervised}.

From those tables, we can see that the LinkDistMLP can outperform the same-structured MLP on almost all 8 datasets by large margins.
While the LinkDistMLP predicts with only the features of the central node, it still matches the GCN in the inductive setting and even exceeds the GCN in the other two settings.
Moreover, utilizing Message Passing (LinkDist) and training contrastively (CoLinkDistMLP, CoLinkDist) can both consistently boost the accuracy.
This indicates the success of distilling self-knowledge from node pairs and that the LinkDists are competitive methods with GNNs in the Node Classification tasks.

We also notice that the GCN2MLP behaves very well in the semi-supervised setting, especially in the inductive setting where there's still a gap between LinkDistMLPs and the GCN2MLP, as Table~\ref{tbl:inductive} and Table~\ref{tbl:transductive} show.
However, in the full-supervised setting as Table~\ref{tbl:full-supervised} shows, the accuracy of the GCN2MLP is almost the same as that of the MLP.
It is far worse than that of the GCN and the LinkDists.
This is because the GCN2MLP in the semi-supervised setting gains additional knowledge from nodes that are outside of the training set, whereas in the full-supervised setting, all nodes are either in the training set or in the evaluation sets.
There is no much additional knowledge to distil.

We tend to experiment with the PPI dataset~\citep{DBLP:conf/nips/HamiltonYL17}, where the training set is formed up with several complete graphs and the trained methods are required to classify nodes on completely unseen graphs, but find that the hyperparameter $\alpha$ in the LinkDist become 0 and LinkDist degenerated into a vanilla MLP\@.
In other words, LinkDist may have problems to induce on entirely unseen graphs.

\section{Conclusions}

In this work, we propose a novel way named LinkDist to classify nodes on graphs without the need of passing messages.

Semi- and full-supervised Node Classification experiments show that LinkDist can predict with comparable accuracy against GNNs even without information from adjacent nodes.

LinkDist can be very practical because it rectifies the three issues of Message Passing:

\begin{itemize}
    \item LinkDist learns in mini-batches of edges.
        It requires less space than using adjacency matrices and can be implemented easily for large graphs.
    \item LinkDist can be deployed to devices with limited resources, predicting fast like MLPs but accurately like GNNs, without the need to store adjacent nodes and their features.
    \item When new nodes or new edges are inserted, LinkDist updates the predictions of existing nodes by optimizing its learnable parameters, instead of propagating information to all nodes.
\end{itemize}

While LinkDist in Message Passing mode (LinkDist and CoLinkDist) can produce very competitive accuracy with GNNs, there's still a gap between LinkDist in non-Message Passing mode (LinkDistMLP and CoLinkDistMLP) and GCN2MLP, if experimenting with the inductive setting.
It shows the power of networks as simple as MLPs and points out the potential of LinkDist.
In the future, we may investigate to find more effective solutions to distil knowledge into MLPs to narrow the gap.

\section*{Broader Impact}
\label{sec:impact}

This work addresses the issues of Message Passing that are widely adopted in Graph Neural Networks.
The proposed LinkDist is a general-purpose method.
It may reduce the cost of mining information from graphic data in existing applications.
To the best of our knowledge, it has no foreseeable negative societal impacts.


{
\small
\bibliographystyle{apalike}
\bibliography{main}
}

\section*{Checklist}

The checklist follows the references.  Please
read the checklist guidelines carefully for information on how to answer these
questions.  For each question, change the default \answerTODO{} to \answerYes{},
\answerNo{}, or \answerNA{}.  You are strongly encouraged to include a {\bf
justification to your answer}, either by referencing the appropriate section of
your paper or providing a brief inline description.  For example:
\begin{itemize}
  \item Did you include the license to the code and datasets? \answerYes{See Section~\ref{gen_inst}.}
  \item Did you include the license to the code and datasets? \answerNo{The code and the data are proprietary.}
  \item Did you include the license to the code and datasets? \answerNA{}
\end{itemize}
Please do not modify the questions and only use the provided macros for your
answers.  Note that the Checklist section does not count towards the page
limit.  In your paper, please delete this instructions block and only keep the
Checklist section heading above along with the questions/answers below.

\begin{enumerate}

\item For all authors...
\begin{enumerate}
  \item Do the main claims made in the abstract and introduction accurately reflect the paper's contributions and scope?
    \answerYes{}
  \item Did you describe the limitations of your work?
    \answerYes{GCN2MLP outperforms our work with inductive setting~\ref{tbl:inductive}. Our work has its problems to induce on entirely unseen graphs~\ref{ssec:results}.}
  \item Did you discuss any potential negative societal impacts of your work?
    \answerYes{See Section~\ref{sec:impact}}
  \item Have you read the ethics review guidelines and ensured that your paper conforms to them?
    \answerYes{}
\end{enumerate}

\item If you are including theoretical results...
\begin{enumerate}
  \item Did you state the full set of assumptions of all theoretical results?
    \answerNA{This work contains no theoretical results.}
	\item Did you include complete proofs of all theoretical results?
    \answerNA{This work contains no theoretical results.}
\end{enumerate}

\item If you ran experiments...
\begin{enumerate}
  \item Did you include the code, data, and instructions needed to reproduce the main experimental results (either in the supplemental material or as a URL)?
  \answerYes{We include the code and instructions enough to reproduce our work in the supplemental material and at the webpage \url{https://github.com/cf020031308/LinkDist}.}
  \item Did you specify all the training details (e.g., data splits, hyperparameters, how they were chosen)?
    \answerYes{Please see Section~\ref{sec:experiment}.}
	\item Did you report error bars (e.g., with respect to the random seed after running experiments multiple times)?
    \answerYes{We include the standard deviations in Table~\ref{tbl:inductive-std}, Table~\ref{tbl:transductive-std}, and Table~\ref{tbl:full-supervised-std} in Appendix~\ref{appendix}.}
	\item Did you include the total amount of compute and the type of resources used (e.g., type of GPUs, internal cluster, or cloud provider)?
    \answerNo{We think the hardwares do not matter. Even with a CPU one can reproduce our work and we use a GPU with 11GB memory.}
\end{enumerate}

\item If you are using existing assets (e.g., code, data, models) or curating/releasing new assets...
\begin{enumerate}
  \item If your work uses existing assets, did you cite the creators?
      \answerYes{They are cited at the beginning of this paper~\ref{sec:introduction} and again at the Section~\ref{sec:experiment} of experiment.}
  \item Did you mention the license of the assets?
    \answerNo{We were unable to find the license for the dataset we used}
  \item Did you include any new assets either in the supplemental material or as a URL?
    \answerYes{I include my code in the supplemental material.}
  \item Did you discuss whether and how consent was obtained from people whose data you're using/curating?
    \answerNA{}
  \item Did you discuss whether the data you are using/curating contains personally identifiable information or offensive content?
    \answerNA{}
\end{enumerate}

\item If you used crowdsourcing or conducted research with human subjects...
\begin{enumerate}
  \item Did you include the full text of instructions given to participants and screenshots, if applicable?
    \answerNA{}
  \item Did you describe any potential participant risks, with links to Institutional Review Board (IRB) approvals, if applicable?
    \answerNA{}
  \item Did you include the estimated hourly wage paid to participants and the total amount spent on participant compensation?
    \answerNA{}
\end{enumerate}

\end{enumerate}


\appendix

\section{Appendix}
\label{appendix}

\begin{table}
  \caption{
      Standard deviations (\%) in the \textbf{semi-supervised} and \textbf{inductive} Node Classification.
      Methods are divided into two groups by whether the method aggregates messages from adjacent nodes in the evaluation phase.
      We bold the highest accuracy for every dataset in each group.}
  \label{tbl:inductive-std}
  \centering
  \setlength{\tabcolsep}{1.4mm}{
  \begin{tabular}{lrrrrrrrr}
    \toprule
      Method & Cora & Citesee & Pubmed & Cora & Amazon & Amazon & Coauthor & Coauthor \\
      & & & & Full & Photo & Computer & CS & Physics \\
    \midrule
MLP & 0.87 & 1.10 & 1.05 & 0.98 & 2.97 & 2.42 & 0.93 & 1.14 \\
GCN2MLP & 0.98 & 0.91 & 1.76 & 1.92 & 2.10 & 1.47 & 0.40 & 0.95 \\
LinkDistMLP & 0.90 & 0.72 & 2.03 & 1.81 & 1.73 & 1.95 & 1.35 & 1.05 \\
CoLinkDistMLP & 0.62 & 1.04 & 0.47 & 1.67 & 1.90 & 2.41 & 0.84 & 0.90 \\
    \midrule
GCN & 0.91 & 0.64 & 1.61 & 1.79 & 1.82 & 2.25 & 1.19 & 1.09 \\
LinkDist & 0.79 & 0.91 & 1.20 & 1.74 & 1.91 & 1.23 & 0.76 & 1.01 \\
CoLinkDist & 0.58 & 0.77 & 0.68 & 2.08 & 2.23 & 2.12 & 0.83 & 1.08\\
    \bottomrule
  \end{tabular}}
\end{table}

\begin{table}
  \caption{
      Standard deviations (\%) in the \textbf{semi-supervised} and \textbf{transductive} Node Classification.
      Methods are divided into two groups by whether the method aggregates messages from adjacent nodes in the evaluation phase.
      We bold the highest accuracy for every dataset in each group.}
  \label{tbl:transductive-std}
  \centering
  \setlength{\tabcolsep}{1.4mm}{
  \begin{tabular}{lrrrrrrrr}
    \toprule
      Method & Cora & Citesee & Pubmed & Cora & Amazon & Amazon & Coauthor & Coauthor \\
      & & & & Full & Photo & Computer & CS & Physics \\
    \midrule
MLP & 0.87 & 1.10 & 1.05 & 0.98 & 2.97 & 2.42 & 0.93 & 1.14 \\
GCN2MLP & 0.83 & 1.74 & 1.68 & 2.14 & 1.23 & 1.73 & 0.89 & 0.79 \\
LinkDistMLP & 0.43 & 0.70 & 1.35 & 1.43 & 2.07 & 2.49 & 1.11 & 1.10 \\
CoLinkDistMLP & 0.61 & 0.79 & 0.38 & 1.83 & 1.88 & 2.05 & 0.68 & 1.65 \\
    \midrule
GCN & 1.12 & 1.24 & 1.39 & 1.68 & 1.53 & 2.55 & 1.79 & 0.86 \\
LinkDist & 0.85 & 0.75 & 0.85 & 2.05 & 1.78 & 2.51 & 1.08 & 1.06 \\
CoLinkDist & 0.46 & 0.74 & 1.02 & 1.43 & 1.88 & 2.52 & 0.98 & 1.42\\
    \bottomrule
  \end{tabular}}
\end{table}

\begin{table}
  \caption{
      Standard deviations (\%) in the \textbf{full-supervised} and \textbf{transductive} Node Classification.
      Methods are divided into two groups by whether the method aggregates messages from adjacent nodes in the evaluation phase.
      We bold the highest accuracy for every dataset in each group.}
  \label{tbl:full-supervised-std}
  \centering
  \setlength{\tabcolsep}{1.4mm}{
  \begin{tabular}{lrrrrrrrr}
    \toprule
      Method & Cora & Citesee & Pubmed & Cora & Amazon & Amazon & Coauthor & Coauthor \\
      & & & & Full & Photo & Computer & CS & Physics \\
    \midrule
MLP           & 1.80 & 2.23 & 0.52 & 0.90 & 0.82 & 0.77 & 0.22 & 0.19 \\
GCN2MLP       & 1.70 & 1.55 & 0.64 & 0.61 & 0.55 & 0.63 & 0.34 & 0.24 \\
LinkDistMLP   & 1.24 & 2.00 & 0.56 & 0.67 & 0.41 & 0.63 & 0.25 & 0.20 \\
CoLinkDistMLP & 1.61 & 1.35 & 0.37 & 0.61 & 0.48 & 0.88 & 0.51 & 0.20 \\
    \midrule
GCN           & 1.67 & 1.59 & 0.82 & 0.76 & 0.64 & 0.54 & 0.45 & 0.23 \\
LinkDist      & 1.23 & 1.90 & 0.44 & 0.66 & 0.30 & 0.51 & 0.32 & 0.22 \\
CoLinkDist    & 1.24 & 1.45 & 0.26 & 0.62 & 0.43 & 0.72 & 0.34 & 0.17 \\
\bottomrule
  \end{tabular}}
\end{table}

\end{document}